\documentclass[machines,article,accept,pdftex,moreauthors]{Definitions/mdpi} 
\usepackage{graphicx}
\usepackage{algorithmic}
\usepackage{algorithm}

\usepackage[labelformat=simple]{subcaption}

\DeclareCaptionLabelFormat{subcaptionlabel}{\normalfont(\textbf{#2}\normalfont)}
\firstpage{1} 
\pubvolume{1}
\issuenum{1}
\articlenumber{0}
\pubyear{2024}
\copyrightyear{2024}
\externaleditor{Academic Editor: Radu-Emil Precup}
\datereceived{18 February 2024} 
\daterevised{21 March 2024} % Comment out if no revised date
\dateaccepted{21 March 2024} 
\datepublished{ } 
\hreflink{https://doi.org/} % If needed use \linebreak
\Title{Discretionary Lane-Change Decision and Control via Parameterized Soft Actor--Critic for Hybrid Action Space}%Attention: titled changed

\TitleCitation{Discretionary Lane-Change Decision and Control via a Parameterized Soft Actor--Critic for Hybrid Action Space}%Attention: titled changed

\Author{Yuan Lin, %MDPI: Please carefully check the accuracy of names and affiliations.
 Xiao Liu * and Zishun Zheng}
\AuthorNames{Yuan Lin, Xiao Liu and Zishun Zheng}

\AuthorCitation{Lin, Y.; Liu, X.; Zheng, Z.}
\address[1]{Shien-Ming Wu School of Intelligent Engineering, South China University of Technology, \mbox{Guangzhou 510641, China}; yuanlin@scut.edu.cn (Y.L.);  202320160023@mail.scut.edu.cn (Z.Z.)\\
	}

\corres{\hangafter=1 \hangindent=1.05em \hspace{-0.82em} Correspondence: 202121060431@mail.scut.edu.cn %MDPI: the QQ email is not allowed, we kept the institutional email
}

\abstract{This study focuses on a crucial task in the field of autonomous driving, autonomous lane change. Autonomous lane change plays a pivotal role in improving traffic flow, alleviating driver burden, and reducing the risk of traffic accidents. However, due to the complexity and uncertainty of lane-change scenarios, the functionality of autonomous lane change still faces challenges. In this research, we conducted autonomous lane-change simulations using both deep reinforcement learning (DRL) and model predictive control (MPC). Specifically, we used the parameterized soft actor--critic (PASAC) algorithm to train a DRL-based lane-change strategy to output both discrete lane-change decisions and continuous longitudinal vehicle acceleration. We also used MPC for lane selection based on the smallest predictive car-following costs for the different lanes. For the first time, we compared the performance of DRL and MPC in the context of lane-change decisions. The simulation results indicated that, under the same reward/cost function and traffic flow, both MPC and PASAC achieved a collision rate of 0\%. PASAC demonstrated a comparable performance to MPC in terms of average rewards/costs and vehicle speeds.}


\begin{document}
\maketitle
	%%%%%%%%%%%%%%%%%%%%%%%%%%%%%%%%%%%%%%%%%%
\setcounter{section}{0} %% Remove this when starting to work on the template.
	%\section{}
%The template details the sections that can be used in a manuscript. Note that the order and names of article sections may differ from the requirements of the journal (e.g., the positioning of the Materials and Methods section). Please check the instructions on the authors' page of the journal to verify the correct order and names. For any questions, please contact the editorial office of the journal or support@mdpi.com. For LaTeX-related questions please contact latex@mdpi.com.%\endnote{This is an endnote.} % To use endnotes, please un-comment \printendnotes below (before References). Only journal Laws uses \footnote.

% The order of the section titles is different for some journals. Please refer to the "Instructions for Authors” on the journal homepage.

\section{Introduction}
The development of autonomous driving has indeed brought revolutionary changes to transportation~\cite{ref1}. Autonomous driving technology not only alleviates the burden on drivers and improves traffic flow but, more importantly, it significantly reduces traffic accidents caused by human errors when driving. According to the World Health Organization, nearly 1.3 million people die in road traffic accidents globally each year, with~94\% attributed to driver errors. In lane-change scenarios in particular, the~actions of surrounding vehicles are often challenging to predict, making automated lane-change a critical task for autonomous~vehicles.

Research has indicated that nearly 10\% of highway accidents are caused by lane-change maneuvers \cite{ref2}. Therefore, a~safe, smooth, and~efficient automated lane-change mechanism is crucial for autonomous vehicles. To~achieve this goal, the~vehicle's architecture must possess efficient and robust execution capabilities that are able to handle uncertainties in the operating environment, make rational decisions, and~execute appropriate actions to cope with the potentially adversarial or cooperative behaviors of surrounding~vehicles.

Currently, automated lane changing in autonomous driving is considered Level 2 automation~\cite{ref3}. Advanced driver-assistance systems such as lane keeping assist (LKA), lane centering control (LCC), and~adaptive cruise control (ACC)~\cite{ref4} are relatively well-established, but~the lane-change function still requires further development and improvement. Although~there has been some progress in research in automated lane-change decision-making, this functionality has not yet been widely implemented in~vehicles.

 MPC stands out as a method of optimizing a sequence of future control actions to address real-time control problems. For~instance, Ji proposed a collision-free trajectory planning method based on artificial potential fields and multi-constraint MPC
~\cite{ref6}. Raffo presented an MPC-based trajectory tracking method, with~two cascaded MPC controllers handling vehicle kinematics and dynamics models, effectively reducing computational complexity~\cite{ref7}. Xu introduced an MPC controller for a lane-keeping system, utilizing a five-point interpolation method to generate a reference trajectory~\cite{ref8}. Similarly, Sameul conducted simulations comparing MPC and PID controllers for trajectory tracking in autonomous vehicles, finding that MPC exhibited better robustness in various scenarios, including vehicle load, longitudinal velocity, and~steering changes~\cite{ref9}. Hang proposed a human-like decision-making framework, combining potential field methods and MPC for collision-free path planning. Additionally, he introduced a module that integrated decision-making and motion planning, considering the social behavior of surrounding traffic participants~\cite{ref10,ref11}.

On the other hand, reinforcement learning (RL) has consistently been a research hotspot in the field of decision-making. For~instance, the~AlphaGo Go-playing robot, which defeated the human Go champion, was a result of training in discrete-action reinforcement learning~\cite{ref12}. Currently recognized discrete-action reinforcement learning algorithms include the Deep Q-Network (DQN)~\cite{ref11}, Double DQN (DDQN)~\cite{ref13}, and Rainbow~\cite{ref14}, among~others. Continuous-action reinforcement learning algorithms include the Deep Deterministic Policy Gradient (DDPG)~\cite{ref15}, Twin-Delayed DDPG (TD3)~\cite{ref16}, Soft Actor Critic (SAC)~\cite{ref17}, and~so~on. 

Few papers have proposed different methods for obtaining hybrid action spaces~\cite{ref18}. One highly cited method is parameterized DDPG (PA-DDPG), introduced in 2016 by scholars from the University of Texas, which utilizes continuous-action reinforcement learning to address hybrid action spaces~\cite{ref19}. Another well-cited method is the Parameterized Deep \mbox{Q-Network} (PDQN), proposed in 2018 by researchers from Tencent AI Lab, which combines actor--critic learning and Q-learning, utilizing Q-learning instead of critic learning in DDPG for discrete action selection~\cite{ref20}. In~2019, scholars from the University of Twente proposed an improved approach called multi-pass P-DQN (MPDQN)~\cite{ref21}, which distributes continuous action inputs to the Q-network based on the correspondence between discrete and corresponding continuous actions, resulting in more reasonable Q-value outputs. In~2022, scholars from Tianjin University introduced the HyAR-TD3 algorithm~\cite{ref22}, which employs representation learning to map continuous action spaces and hybrid action~spaces.

Currently, most literature uses discrete reinforcement learning to achieve optimal control for non-mandatory automated lane changing of autonomous vehicles~\cite{ref23,ref24, ref25,ref26,ref27}. Typically, these papers adopt a hierarchical control approach, where the upper-level control outputs lane-change decisions using discrete reinforcement learning (discrete control variables), and~the lower-level control uses a car-following model to output the vehicle acceleration (continuous control variables). However, only a few studies have applied hybrid-action reinforcement learning to automated lane-change decision-making and control. In~2021, scholars from the University of Washington proposed the Hybrid Deep Q-Learning and Policy Gradient (HDQPG) to achieve automated lane-change of vehicles~\cite{ref28}.

In this paper, we use the PASAC algorithm tailored for hybrid-action spaces. We trained the model using traffic simulation software on the SUMO platform for various traffic scenarios. To~validate the algorithm's superior performance in terms of stability and optimality, we compared the results of PASAC with MPC, considering metrics such as collision rate, average speed, value function, and~jerk. However, it is crucial to note some known differences between the two approaches. First, MPC requires online optimization and demands relatively powerful computing resources for real-time applications, raising monetary concerns about practical deployment~\cite{ref29}. On~the other hand, the~DRL solution, based on neural networks, despite being time-consuming during offline training, has short execution times and is suitable for real-time applications. Second, MPC relies on a model-based approach, while DRL control solutions, based on black-box neural networks, lack theoretical guarantees~\cite{ref30}. 

In our MPC model, the~ego vehicle needs to assess whether executing a lane-change maneuver is beneficial. If~deemed beneficial, it adjusts its position and speed to prepare for the lane change; otherwise, it chooses to follow the preceding vehicle. In~RL, the~intelligent agent interacts with the environment, selects actions based on the current state, and~continually updates its policy based on environmental feedback in the form of rewards. The~intelligent agent in reinforcement learning can learn adaptive driving strategies for lane-change problems, enabling vehicles to make intelligent lane-change~decisions.

The primary contribution of this work is the introduction of the PASAC algorithm for discretionary lane changing, as well as~the first quantitative and comprehensive comparison of the hybrid-action space reinforcement learning algorithm PASAC with MPC. We experimentally verified the superiority of PASAC in lane-change decision and control, and conducted a detailed analysis of its performance. To~the best of our knowledge, such a comparison does not exist in the literature. This not only provides new insights into the application of hybrid-action space reinforcement learning in practical control problems but also offers empirical support for the comparison of reinforcement learning with traditional control~methods. 

Regarding the structure of this paper, the~second section provides a detailed introduction to the PASAC algorithm. The~application of PASAC and MPC in lane-change scenarios is discussed in the third and fourth sections, respectively. The~fifth section compares the DRL and MPC methods. Finally, conclusions are drawn in the sixth~section.

%\begin{quote}
%This is an example of a quote.
%\end{quote}

%%%%%%%%%%%%%%%%%%%%%%%%%%%%%%%%%%%%%%%%%%
\section{Parameterized Soft~Actor--Critic}
In this section, we present an overview of the hybrid action space structure using the SAC~algorithm.

\subsection{Reinforcement~Learning}
Reinforcement learning is a learning method employed for decision-making and control. In~reinforcement learning, an~agent takes actions based on the current time step's environmental state, and~subsequently, the~environment transitions to a new state in the next time step as a result of that action. The~agent also receives rewards based on the actions taken, and~both the actions and rewards have a certain probabilistic nature. The~objective of reinforcement learning algorithms is to learn effective policies by maximizing the expected discounted cumulative reward for each episode. Specifically, the~discounted cumulative reward for a state-action pair is referred to as the Q-value, denoted as%MDPI: Please confirm if all bolds in equations are necessary or not. we just highlight here.
 $ Q(\textbf s_{t},\textbf a_{t}) = \mathbb {E}\left [^{\tau=T}_{\tau = t} {\gamma}^{\tau-T} r({\textbf s_{\tau},\textbf a_{\tau}} ) \right]$. Here, $r({\textbf s_{\tau},\textbf a_{\tau}} )$
represents the reward for the state $s$ and action $a$ at time step $\tau$, and~$\gamma \in$[0,1] is the discount factor. The~resolution of reinforcement learning problems adheres to the Bellman optimality principle. This principle asserts that if the optimal Q-value for the next step is known, then the action for the current time step must also be optimal. In~other words, for~an optimal policy, $Q(\textbf s_t,\textbf a_t)^* =r(\textbf s_t,\textbf a_t)+\gamma {Q}^*(\textbf s_{t+1},\textbf a_{t+1}) $, with~$*$ denoting the optimality. This principle forms the foundation for devising effective policies in reinforcement~learning.

\subsection{Soft~Actor--Critic}
The actor--critic architecture is a core component of the RL algorithm, as~proposed by Sutton and Barto (1999)~\cite{ref31}. It is used to solve action selection and value function learning. In~this context, we consider a parameterized state value function $V$, a~soft Q function, and~a policy network. The~parameters of these networks are denoted as $\psi$, $\hat\psi$, $\theta$, and $\phi$, respectively. The~SAC algorithm considers the maximum entropy objective in reinforcement learning's maximum expectation, and~the modified expectation is
\begin{equation}
	J(\pi) = \sum_{t=0}^{T} \mathbb{E}_{(\textbf s_t, \textbf a_t) \sim \rho_\pi} \gamma^t \left[ r(\textbf s_t, \textbf a_t) + \alpha H(\pi(\cdot |\textbf s_t)) \right],
\end{equation}
In this formula, “$\cdot$” represents all possible actions. $\rho_\pi$ denotes the new policy. The~higher the entropy $H$, the~stronger the system's uncertainty. In~other words, a~policy with higher entropy provides more significant action unpredictability. To~regulate the impact of entropy on the policy, the~SAC algorithm introduces a hyperparameter $\alpha$, which plays a pivotal role in determining the relative significance of the entropy term on rewards.
\begin{equation}
	\pi^*  =\arg \underset{\pi}{\max}\,\mathbb{E}\underset{\tau \sim \pi}{\phantom{.}}\left[ \sum_{t=0}^{\infty}\gamma^t(R(\textbf s_t,\textbf a_t,\textbf s_{t+1})) + \alpha H(\pi(\cdot | \textbf s_t))\right],
\end{equation}
The primary objective in training the soft value function is to minimize the square of residuals. In~essence, through the optimization of the soft value function, the~goal is to diminish the disparity between model predictions and actual observations, thereby enhancing the overall efficacy of the training process.
\begin{equation}
	J_V(\psi) = \mathbb{E}_{{s_t} \sim D} \left[ \frac{1}{2} (V_{\psi} (\textbf s_t) -E_{\textbf a_t \sim \pi_\psi}[Q_\theta(\textbf s_t,\textbf a_t)-log\pi_\phi(\textbf a_t|\textbf s_t)])^2  \right],
\end{equation}
The gradient can be estimated using an unbiased estimator
\begin{equation}
	\hat {\bigtriangledown}_\psi J_V(\psi) ={\bigtriangledown}_\psi V_{\psi} (\textbf s_t)(V_{\psi} (\textbf s_t)-Q_\theta(\textbf s_t,\textbf a_t)+log\pi_\phi(\textbf a_t|\textbf s_t))  ,
\end{equation}
The parameters of the soft Q-value function are determined by minimizing the residual of the Bellman equation.
\begin{equation}
	J_Q(\theta) = \mathbb{E}_{{( \textbf s_t,\textbf a_t )} \sim D} \left[ \frac{1}{2} \left(Q_\theta(\textbf s_t,\textbf a_t)-\hat Q(\textbf s_t,\textbf a_t) \right) ^2 \right],
\end{equation}
with
\begin{equation}
\hat Q(\textbf s_t,\textbf a_t) =r(\textbf s_t,\textbf a_t)+\gamma \mathbb{E}_{{s_t} \sim p} \left [V_{\hat \psi} (s_{t+1})          \right],
\end{equation}
After optimizing with a stochastic gradient
\begin{equation}
	\hat {\bigtriangledown}_\theta J_Q(\theta) ={\bigtriangledown}_\theta Q_{\theta} (\textbf s_t,\textbf a_t)(Q_{\theta} (\textbf s_t,\textbf a_t)-r(\textbf s_t,\textbf a_t)-\gamma V_{\hat \psi} (s_{t+1})),
\end{equation}
The method of translating strategies using neural networks is as follows:
\begin{equation}
	\textbf a_t =f_{\phi}(\epsilon_t;\textbf s_t),
\end{equation}
$\epsilon_t$ represents an input noise vector sampled from a fixed distribution, such as a spherical Gaussian distribution.The objective function is denoted as
\begin{equation}
	J_{\pi}(\phi) = \mathbb{E}_{{s_t} \sim D,\epsilon_t \sim N} \left[ log\pi_\phi(f_\phi(\epsilon_t;\textbf s_t)|s_t) -Q_\theta(\textbf s_t,f_\phi(\epsilon_t;\textbf s_t)) \right],
\end{equation}
where $\pi_\phi$ is defined by the function $f_\phi$, the~gradient of Equation~(9) is as follows:
\begin{equation}
	\hat {\bigtriangledown}_\phi J_{\pi}(\phi) ={\bigtriangledown}_\phi log\pi_\phi(\textbf a_t|\textbf s_t)+({\bigtriangledown}_{\textbf a_t}log\pi_\phi(\textbf a_t|\textbf s_t)-{\bigtriangledown}_{\textbf a_t}Q(\textbf a_t,\textbf s_t)){\bigtriangledown}_\phi f_\phi(\epsilon;\textbf s_t).
\end{equation}
\subsection{Parameterized Soft~Actor--Critic}
In this context, we define a Markov decision process with a parameterized action space. The~action space consists of a set of discrete actions, denoted as \linebreak $A_d = {a_1, a_2, ...... a_n}$. Each discrete action $a \in A_d$ is associated with a corresponding set of continuous parameters, represented as ${a^{p_1}_1, a^{p_2}_2, ...... a^{p_n}_n}$. In~our environment, the~actor network outputs $m$ continuous parameters to form continuous actions and selects $n-m$ continuous parameters as weights for the discrete actions ($m<n$). The~discrete action is determined by choosing the action with the maximum weight among the $n-m$ continuous parameters, expressed as $a_d = \max(a_{m+1}, a_{m+2}, ...... a_n)$. The~role of the actor network is to simultaneously decide which discrete action to execute and how to parameterize that action. Here, we adopt an approach similar to Delalleau 2019~\cite{ref32}, but~unlike the former, our discrete actions are deterministic rather than~stochastic.

The PASAC algorithm is similar to the algorithm proposed by Peter Stone in 2016~\cite{ref19}, as~illustrated below: The actor neural network can directly output continuous actions, and~for discrete actions, it outputs the action with the maximum weight, where the weights are normalized within the range [0, 1]. The training process of the PASAC algorithm is shown in Algorithm \ref{alg1}.

%\begin{table}[H]
%\setlength{\arrayrulewidth}{2pt}

\begin{algorithm}[H]
\caption { PASAC Algorithm Training Process.}
\label{alg1}
\begin{algorithmic}
\STATE  \hspace{0cm} \textbf{Input:} $\theta$, $\psi$, $\hat{\psi}$, $\phi$ 
\STATE \hspace{2cm}$\hat{\psi}$ $\gets$ $\psi$, $D$ $\gets$ $\emptyset$
\STATE \hspace{1cm} \textbf{For} each iteration \textbf{do}
\STATE \hspace{1.5cm} \textbf{For} each environment step \textbf{do}
\STATE \hspace{1.9cm}  $(\textbf a_c,\textbf k_d) \sim  \pi_{\phi}(\textbf a_t|\textbf s_t)$
\STATE \hspace{2cm} $\textbf a_d \sim  argmax \ \textbf k_d$
\STATE \hspace{2cm} $\textbf s_{t+1} \sim p(\textbf s_{t+1}|\textbf s_t,\textbf a_t) $
\STATE \hspace{1.95cm} $D \gets D \bigcup \{(\textbf s_t,\textbf a_t,\textbf r(\textbf s_t,\textbf a_t),\textbf s_{t+1})\}$
\STATE \hspace{1.5cm} \textbf{End for} 
\STATE \hspace{1.5cm} \textbf{For} each gradient step \textbf{do}
\STATE \hspace{2cm} ${\theta_i}\gets \theta_i-\lambda_Q \hat {\bigtriangledown}_{\theta_i}J_Q(\theta_i) \ $for$\ i \in\{ 1,2\}$
\STATE \hspace{2cm} ${\phi}\gets \phi-\lambda_\pi \hat {\bigtriangledown}_{\phi}J_\pi(\phi) $
\STATE \hspace{2cm} ${\psi}\gets \psi-\lambda_V \hat {\bigtriangledown}_{\psi}J_V(\psi) $
\STATE \hspace{2cm} $\hat{\psi_i}\gets \tau  \psi_i + (1-\tau)\hat \psi_i \ $
\STATE \hspace{1.5cm} \textbf{End for} 
\STATE \hspace{1.0cm} \textbf{End for} 
\STATE \hspace{0cm} \textbf{Output} $\theta$, $\psi$, $\phi$ 
\end{algorithmic}
\end{algorithm} 

\begin{figure}[H]
	\begin{subfigure}{0.5\textwidth}
%		\centering
		\includegraphics[width=5.5 cm]{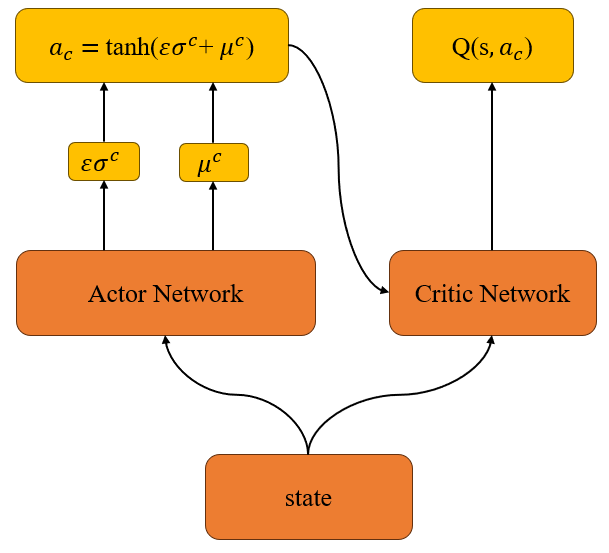}
		\caption{\centering Standard~SAC}
		\label{fig:subfig-a}
	\end{subfigure}
	\begin{subfigure}{0.5\textwidth}
%		\centering
		\includegraphics[width=5.5 cm]{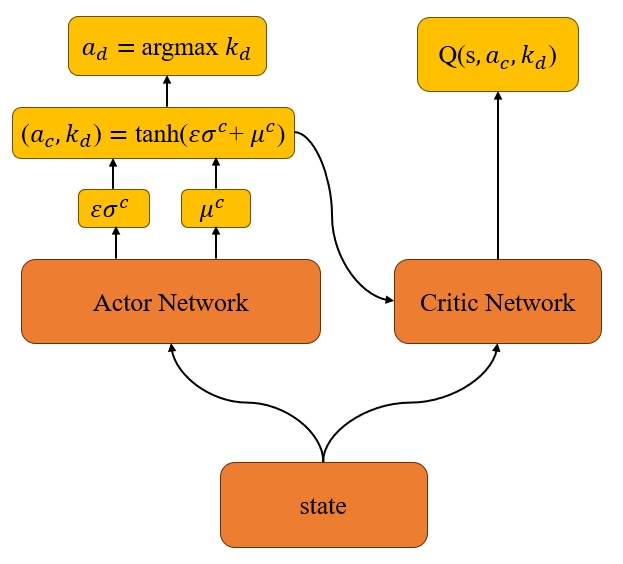}
		\caption{\centering~PASAC}
		\label{fig:subfig-b}
	\end{subfigure}
	\label{fig:two-images}
\caption{ (\textbf{a}) The framework on the left is the standard SAC architecture designed for continuous operation. The~actor outputs mean and standard deviation vectors $\mu$ and $\sigma$, which are utilized for injecting standard normal noise $\epsilon$ and applying the tanh nonlinearity (to keep the actions within a bounded range). The~critic estimates the corresponding Q value based on the state and the actor's action $a_c$. (\textbf{b}) On the right, we use the parameterized SAC structure. including the mean $\mu$ and the variance $\sigma$ for the continuous components. It outputs continuous actions $a_c$ and $k_d$. The~largest $k_d$ among continuous actions is selected for the discrete action. The~critic network still takes the state $s$, continuous actions $a_c$ and $k_d$ as~inputs.\label{fig1}}
\end{figure}

%\end{table}
In order to gain a comprehensive understanding of the decision-making process of the PASAC algorithm, we provide a detailed explanation of its neural network framework. Refer to Figure~\ref{fig1} for an illustration. In~the structure diagram, the~agent has two branches for handling actions---one for processing continuous actions and another for processing discrete actions. The~outputs of these two branches are integrated into the final action decision, enabling the agent to learn and execute tasks in a mixed-action~space.

\vspace{-6pt}

\section{PASAC for Lane~Changing}
In this section, we utilize the open-source simulator SUMO~\cite{ref33}. This integrated framework is employed to construct the RL environment, governing the behavior of autonomous vehicles. We present a model for autonomous lane changing based on reinforcement learning. By~explicitly modeling states, actions, and~rewards, the~objective is to realize intelligent lane-change decisions for vehicles navigating through intricate traffic~scenarios.

\subsection{Scenario~Settings}
\begin{figure}[H]
%	\centering
	\includegraphics[width=10.5 cm]{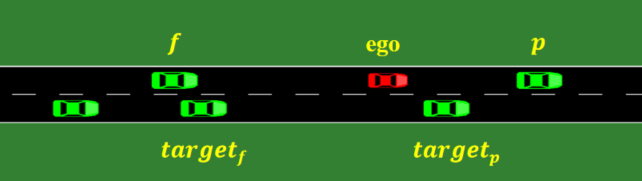}
	\caption{Lane change scenario in~SUMO. \label{fig2}}
\end{figure}

In the conducted experiments detailed in this paper, we utilized a straight roadway with a length of 1000 m and two lanes. The~lane change scenario in SUMO is shown in Figure~\ref{fig2}, the~red car represents the ego vehicle, while the green cars represent the surrounding vehicles~.
 
\subsection{State}
At time $t$, the~distance between the ego vehicle and the preceding vehicle $d^p_t$, the~distance between the ego vehicle and the following vehicle $d^f_t$, the~distance from the ego vehicle to the preceding vehicle in the target lane $d^{target_p}_t$, the~distance from the ego vehicle to the following vehicle in the target lane $d^{target_f}_t$, ego vehicle’s speed $v_{ego}$, ego vehicle’s acceleration $a^{ego}_t$, the~speeds of the preceding car and following car $v^p_t,v^f_t$, as~well as the speeds of the preceding and trailing cars in the target lane $v^{target_p}_t,v^{target_f}_t$.
\begin{equation}
	s=(d^p_t,d^f_t,d^{target_p}_t,d^{target_f}_t,v^{target_p}_t,v^{target_f}_t,v^{ego}_t,a^{ego}_t,v^p_t,v^f_t)\in S
\end{equation}
\subsection{Action}
 In this context, we define the action space as
\begin{equation}
	a = \{{a^{ego}_t,0,1}\} \in A,
\end{equation}
In Equation~(12), the~symbol ‘0' signifies the choice to postpone the lane change, indicating the intent to maintain the current position within the ego lane. Conversely, the~symbol ‘1' represents an immediate decision to execute the lane change, manifesting the intention to promptly transition to the target lane. These symbols denote discrete actions. ‘$a^{ego}_t$', on~the other hand, is a continuous action representing the acceleration of the ego~vehicle.
\subsection{Reward}
The reward function is crafted with the objective of motivating positive behaviors and discouraging undesirable actions within the decision-making process of autonomous vehicles. In~this paper, distinct rewards are allocated for tasks such as distance control, successful lane changes, adherence to speed limits, and~collision avoidance. Drawing upon this concept, we formulated the following reward function.
\begin{equation}
	R_{total} = R_{act}+R_{act1}+R_{act2}+R_{collision}
\end{equation}
where $R_{total}$ is the total reward for the simulation scene. Where $R_{collision}$is the penalty for vehicle collisions.
\begin{equation}
	R_{act} = -\omega_0 \lvert y_{t-1} -y_t\rvert
\end{equation}
In Equation~(14), $R_{act}$ is the penalty for frequent lane changes by vehicles, $w_0$ is the corresponding weight, and $y_t$ and $y_{t-1}$ represent the current and previous time step's lateral positions of the vehicle.
Note that when the distance to the preceding vehicle satisfies the ACC spacing, this reward penalty will not be computed, thus aligning with the MPC cost~function.

Through a comprehensive analysis of the disparity between the actual speed and the desired speed of the ego vehicle, coupled with meticulous management of the spacing between the ego vehicle and its preceding and following counterparts, we skillfully crafted a longitudinal acceleration control strategy. The~primary aim of this strategy is to mitigate the likelihood of collisions between vehicles, strategically initiating lane-change maneuvers during instances of reduced speed in the ego vehicle, thereby further optimizing the overall travel time. In~consideration of passenger comfort, we implemented a penalty mechanism for changes in longitudinal acceleration, seeking to strike a harmonious balance between driving efficiency and the overall passenger experience. The~specific reward function is delineated as follows:
\begin{equation}
	R_{act1} = -\omega_1 \lvert d^p_t -d_{safe}\rvert-\omega_2 \lvert d^f_t -d_{safe}\rvert-\omega_3 \lvert v_{ego} -v_{safe}\rvert
\end{equation}
\begin{equation}
	R_{act2} = -\omega_4 \lvert jerk\rvert
\end{equation}
In Equations~(15) and (16), $w_1$, $w_2$, $w_3$, and~$w_4$ denote the corresponding weights. Here, $d_{\text{safe}}$ represents the desired safe distance, $v_{\text{safe}}$ is the desired safe speed, and~$jerk$ signifies the rate of change of acceleration for the ego~vehicle.

In order to comprehensively present the key parameters involved in our analysis, we introduce a parameter table (Table \ref{tab1}) at this point. It is worth noting that the weights were determined though manual~turning. 
\begin{table}[H] 
	\caption{Simulation~parameters.\label{tab1}}
	\newcolumntype{C}{>{\centering\arraybackslash}X}
	\begin{tabularx}{\textwidth}{CCCC}
		\toprule
		\textbf{Parameters}	& \textbf{Value}	& \textbf{Weights} & \textbf{Value}  \\
		\midrule
		$a_{min}$		&$-$4.5~m/s%MDPI: we removed italic of `s', please confirm. The following highlights are the same.
$^2$		& $w_0$   &3.13\\
		$a_{max}$		& 2.6~m/s$^2$		& $w_1$   &0.5\\
		$v_{safe}$	& 13.89~m/s			& $w_2$       &0.4\\
		$d_{safe}$		& 25~m		& $w_3$   &0.72\\
		$R_{collision}$		& $-$200		    & $w_4$       &0.5\\
		\bottomrule
	\end{tabularx}
	%\noindent{\footnotesize{\textsuperscript{1} Tables may have a footer.}}
\end{table}
\unskip

\section{ Model Predictive Control~Model}
In MPC, the~control inputs are determined by solving an optimization problem at each time step, taking into account the current state of the system and predicting its evolution over the horizon. This optimization process aims to minimize a predefined cost function, Here, we compare the costs for different lanes and initiate a lane change for the lane with the lowest cost. The~lane change is instantaneous, wherein no lateral control is considered, the same as for~DRL.

It is worth noting that we used YALMIP to handle optimization solutions. By~leveraging the open-source YALMIP, we formulated and solved the MPC optimization problem. YALMIP can be installed in MATLAB, providing programmers with various shooting and optimization methods to address nonlinear optimization problems.
In this section, the~principles of the decision control for the self-driving vehicle under MPC are introduced. These include the state-space equations, cost function, constraints, future state estimation, and~variable-spacing~strategy.

\subsection{State-Space~Equations}
 The state-space equations for the MPC we implemented are as follows: It is worth noting that we simplified vehicles to a point mass, without~considering the vehicle dynamics~\cite{ref34}, as for DRL.
\begin{equation}
	\begin{aligned}
		d^p_t &= s_p - s, \\
		d^f_t&= s - s_f, \\
		v_{ego} &=\dot{s},\\
		\triangle v^{p}_t &= \dot{s}_p - \dot{s} = v_p - v_{ego}, \\
		\triangle v^{f}_t &= \dot{s} - \dot{s}_f = v_{ego} - v_f, \\
		a^{ego}_t &= \dot{v}_{ego}, \\
    	j^{ego}_t &= \dot{a}^{{ego}_t}
	\end{aligned}
\end{equation}
\begin{equation}
x=\left[d^p_t,d^f_t,v_{ego},\triangle v^{p}_t,\triangle v^{f}_t,a^{ego}_t,j^{ego}_t \right]^T
\end{equation}
\begin{equation}
u = a
\end{equation}
\begin{equation}
x(k+1)=\mathbf{A}x(k)+\mathbf{B}u(k)
\end{equation}
with
\begin{equation}
	\mathbf{A}=
	\begin{bmatrix}
		1 & 0 &-T_s& T_s & 0& 0 &0 \\
		0 & 1 & T_s& 0 & -T_s& 0 & 0 \\
		0 & 0 & 1& 0 & 0& T_s& 0\\
		0 & 0 & 0& 1 & 0& 0 & 0\\
		0 & 0 & 0& 0 & 1&0& 0\\
		0 & 0 & 0& 0 & 0& 0 & 0\\
		0 & 0 & 0& 0 & 0& -1/T_s & 0
	\end{bmatrix},
	\mathbf{B}=
   \begin{bmatrix}
	0  \\
	0 \\
	0 \\
	0\\
	0\\
	1\\
	1/T_s
   \end{bmatrix},
\end{equation}
In Formula, $s$ represents the longitudinal coordinate of the vehicle. $x$ is utilized as the input, where $d^p_t$and $d^f_t$ denote the distances to the preceding and following vehicles, respectively. The~variables $\triangle v^{p}_t$ and $\triangle v^{f}_t$ represent the ego vehicle velocity differences with the preceding and following vehicles, while $v_{ego}$ , $a^{ego}_t$ , and~$j^{ego}_t$denote the velocity, acceleration, and~jerk of the vehicle, respectively.
$u$ represents the control variable, where $T_s$ is the time interval, with~$T_s$ set to 0.1~s.

\subsection{Cost~Function}
\begin{equation}
	J = \omega_1 \lvert d^p_t -d_{safe}\rvert+\omega_2 \lvert d^f_t -d_{safe}\rvert+\omega_3 \lvert v_{ego} -v_{safe}\rvert+\omega_4 \lvert jerk\rvert
\end{equation}
Formula (22) is consistent with the PASAC reward function. The~primary objectives of the first and second terms are to ensure appropriate distances with the lead and following vehicles. The~distance with the following vehicle is not penalized for the current lane, while it is penalized for the target lane. The~third term aims to maintain a safe ego speed. The~fourth term is designed to enhance driving comfort by penalizing jerk. The~MPC cost does not include a penalty for frequent lane~changes.

\subsection{Future State~Estimation}
In the prediction horizon of MPC, scholars like Paolo Falcone fixed the values of slip and friction coefficients within the predictive time horizon, ensuring they remained constant and equal to the estimated values at the current moment~\cite{ref35}. Similarly, in~this paper, we chose N = 5 as the prediction horizon and utilized the same velocity of the leading vehicle at the current time step during the prediction~horizon.  
\subsection{TLACC~(Two-Lane Adaptive Cruise Control)}
TLACC is a decision and control {algorithm}~based on~MPC (Algorithm \ref{alg2}). Where $J_c$ and $J$ denote the future driving costs for the current lane and the target lane, respectively. $l_{sw}$ is the lane change signal, where 0 indicates staying in the current lane, and~1 indicates a change to the target lane. $J_{th}$ is the threshold value for the future driving cost on the current lane that must be satisfied for a lane change, with$ J_{th}$ set to 0.8. $u^c_d$ is the desired control input, and~$u_d$ and $u_{target}$ are the expected control inputs for driving on the current lane and the target lane, respectively. $k_p$ is the extra weight for the future driving cost during a lane change, with~$k_p$ set to 0.1. The~extra weight prevents frequent lane~changes.

%\begin{table}[H]
%	\setlength{\arrayrulewidth}{2pt}
	\begin{algorithm}[H]
		\caption {TLACC %MDPI: Please cite the algorithm in the text and ensure the first citation of each algorithm appears in numerical order. Please check above where we suggested.
 Algorithm Process.}
		\label{alg2}
		\begin{algorithmic}
			\STATE  \hspace{0cm} \textbf{Input:}  	$d^p_t,d^f_t,\triangle v^{p}_t,\triangle v^{f}_t,a^{ego}_t,j^{ego}_t,l^c_t ,v_{ego}$
			\STATE  \hspace{0cm} \textbf{Output:}  	$u_d,l_{sw}$
			\STATE \hspace{1cm} \textbf{While} TLACC engaged \textbf{do}
			\STATE \hspace{1.5cm} $(J_c,u^c_d)\gets MPC(d^p_t,d^f_t,\triangle v^{p}_t,\triangle v^{f}_t,a^{ego}_t,j^{ego}_t,l^c_t,v_{ego})$
			\STATE \hspace{1.5cm} \textbf{If $J_c \leq J_{th}$} \textbf{Then}
			\STATE \hspace{2.5cm} \textbf{Return }$l_{sw} \gets 0,u_d\gets u^c_d$ 
			\STATE \hspace{1.5cm} \textbf{else}
			\STATE \hspace{2.5cm}$(J,u_{target})\gets MPC(d^p_{target},d^f_{target},\triangle v^{p}_{target},\triangle v^{f}_{target},a^{ego}_t,j^{ego}_t,l_{target},v_{ego})$
			\STATE \hspace{2.5cm} \textbf{If}$(1+k_p)J\leq J_c$ \textbf{Then}
			\STATE \hspace{3.5cm} \textbf{Return }$l_{sw} \gets 1,u_d\gets u_{target}$ \STATE \hspace{2.5cm} \textbf{else}
			\STATE \hspace{3.5cm} \textbf{Return }$l_{sw} \gets 0,u_d\gets u^c_d$ 
			\STATE \hspace{2.5cm} \textbf{End } 
			\STATE \hspace{1.5cm} \textbf{End } 
			\STATE \hspace{1.0cm} \textbf{End } 
		\end{algorithmic}
	\end{algorithm} 
%\end{table}

\section {Comparison Results of DRL and MPC }
Under the same conditions of relevant cost functions, input states, and~traffic flow, this section presents the test results of the DRL and MPC~controllers.

\subsection{DRL~Training}
For the training of the reinforcement learning model, we chose total simulation timesteps of 300,000, with~each timestep set to 0.1 s. \textls[-15]{The~training was conducted on a computer equipped with an 8-core (16-thread) AMD processor and an NVIDIA GeForce RTX 3050 Ti GPU, and~the training process took approximately 3 h. It is noteworthy that, before~the start of each episode, there was a 50-m buffer for the initialization of main-road~traffic.}

In our study, to~achieve effective training of the reinforcement learning model, we meticulously selected and configured a set of crucial hyperparameters. The~choice of these hyperparameters directly impacted the model's performance and the stability of the training process. In~Table~\ref{tab2}, we provide a detailed list of the hyperparameters utilized during training, along with their corresponding~values.
\begin{table}[H] 
	\caption{PASAC~Hyperparameters.\label{tab2}}
	\newcolumntype{C}{>{\centering\arraybackslash}X}
	\begin{tabularx}{\textwidth}{CCCC}
		\toprule
		\textbf{Hyperparameters}	& \textbf{Value}	& \textbf{Hyperparameters} & \textbf{Value}  \\
		\midrule
		Discount factor		&0.99		& Tau   &0.005\\
	    Alpha 		& 0.05		&Learning starts   &500\\
		Actor learning rate		& 0.0001		& Mini-batch size   &128\\
		Critic learning rate	& 0.001			& Buffer size   &10,000    \\
		\bottomrule
	\end{tabularx}
	%\noindent{\footnotesize{\textsuperscript{1} Tables may have a footer.}}
\end{table}
\unskip
\subsection{DRL~Testing}
The trained policy underwent additional testing with an extended 350,000 simulation time steps, representing 500 episodes. In~order to better assess the performance of the model, we selected a typical initial condition where the leading vehicle's initial velocity was set to 12.89 m/s, and~the ego vehicle's initial velocity was set to 13.89 m/s. The~traffic flow density was 0.11 vehicles/second, for~two~lanes.
\subsection{Comparison and~Analysis}
 In the following sections, we compare the performance of MPC and RL in executing lane-change tasks for autonomous~vehicles.

\begin{figure}[H]
%\hspace{1.6cm} 
%\raggedright
\includegraphics[width=9 cm]{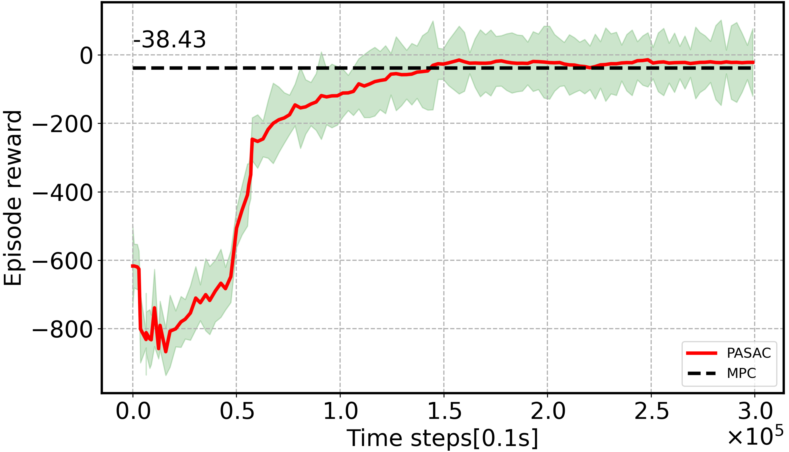}
\caption{The reward(cost) between MPC and~PASAC.\label{fig3}}
\end{figure}

In Figure~\ref{fig3}, the~solid line represents the training curve of PASAC, and~it can be observed that it approached convergence around 150,000 steps. The~dashed line represents the total cost of MPC averaged over 5 episodes. The~performance comparison results between PASAC and MPC are shown in Table~\ref{tab3}. The~average speed and cost for PASAC were superior to MPC. This implies that PASAC achieved better speed and time performance. Additionally, PASAC tended to execute more lane changes compared to~MPC.

\vspace{-9pt}
\begin{table}[H] 
	\caption{Comparison of results from 100 episodes of~testing.\label{tab3}}
	\newcolumntype{C}{>{\centering\arraybackslash}X}
	\begin{tabularx}{\textwidth}{cCCCCC}
		\toprule
		\textbf{}	& \textbf{Collision}	& \textbf{Average Speed (m/s)}&\textbf {Lane Change Times} & \textbf {Reward (Cost)} &\textbf{Reward (Cost) Difference} \\
		\midrule
		PASAC	&0\%		& 14.34 &34  &$-$27.73&27.90\%\\
		MPC		& 0\%		&13.95 &25  &$-$38.46&0\% \\
		\bottomrule
	\end{tabularx}
	%\noindent{\footnotesize{\textsuperscript{1} Tables may have a footer.}}
\end{table}

\begin{figure}[H]
%	\hspace{2cm}
%	\centering
%	\raggedright
	\includegraphics[width=7cm]{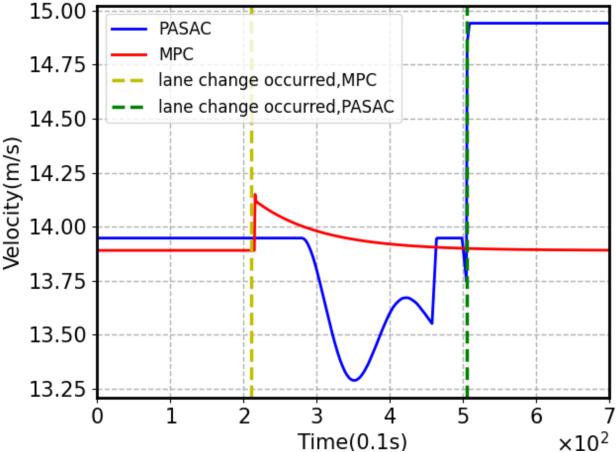}
	\caption{Lane change in the~simulation. \label{fig4}}
\end{figure}

In Figure~\ref{fig4}, the~red solid line represents the self-driving vehicle controlled by the MPC method, while the deep blue solid line represents the vehicle controlled by the reinforcement learning algorithm PASAC. With~PASAC, the~vehicle decelerated suddenly and then accelerated, while with MPC, it accelerated first and then decelerated. The~green and yellow dashed lines respectively represent lane changes by the self-driving vehicle under PASAC and~MPC.

It can be observed that the lane changes under PASAC and MPC occurred at different times (MPC occurred around 22~s, while PASAC occurred around 51~s). This was because, for~PASAC and MPC, the~surrounding vehicles were in different states before the sudden lane change occurred. The~self-driving vehicle under PASAC changed lanes immediately after a sudden deceleration, then accelerated to maintain a higher speed. On~the other hand, the~self-driving vehicle under MPC changed lanes after a sudden acceleration, maintaining a stable~speed.

\begin{figure}[H]
%	\centering
	\includegraphics[width=0.5\textwidth]{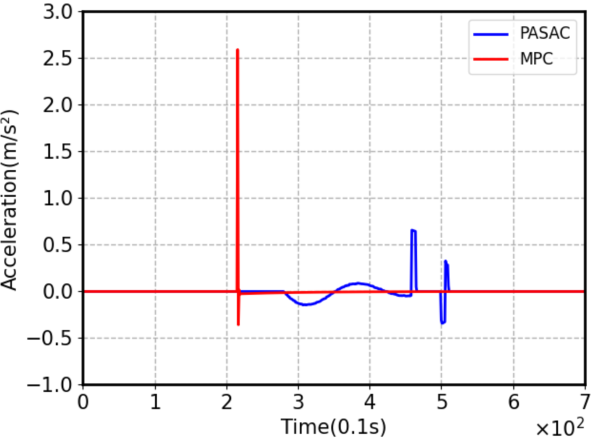} % 更换为图片1的路径
	\includegraphics[width=0.49
\textwidth]{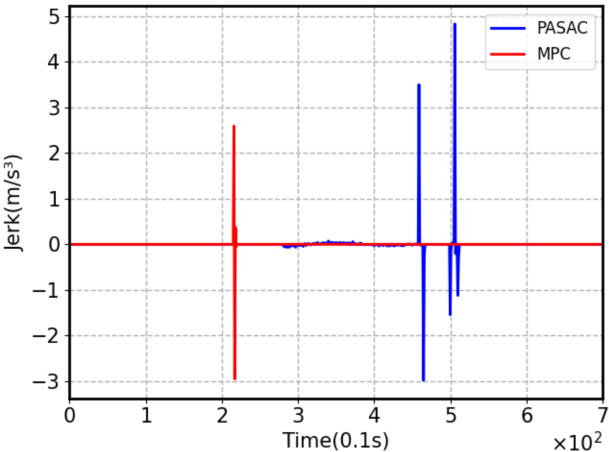} % 更换为图片2的路径
	\caption{Acceleration and jerk during the~simulation.\label{fig5}}
	\label{fig:combined}
\end{figure}

\begin{figure}[H]
%	\centering
	\includegraphics[width=0.48\textwidth]{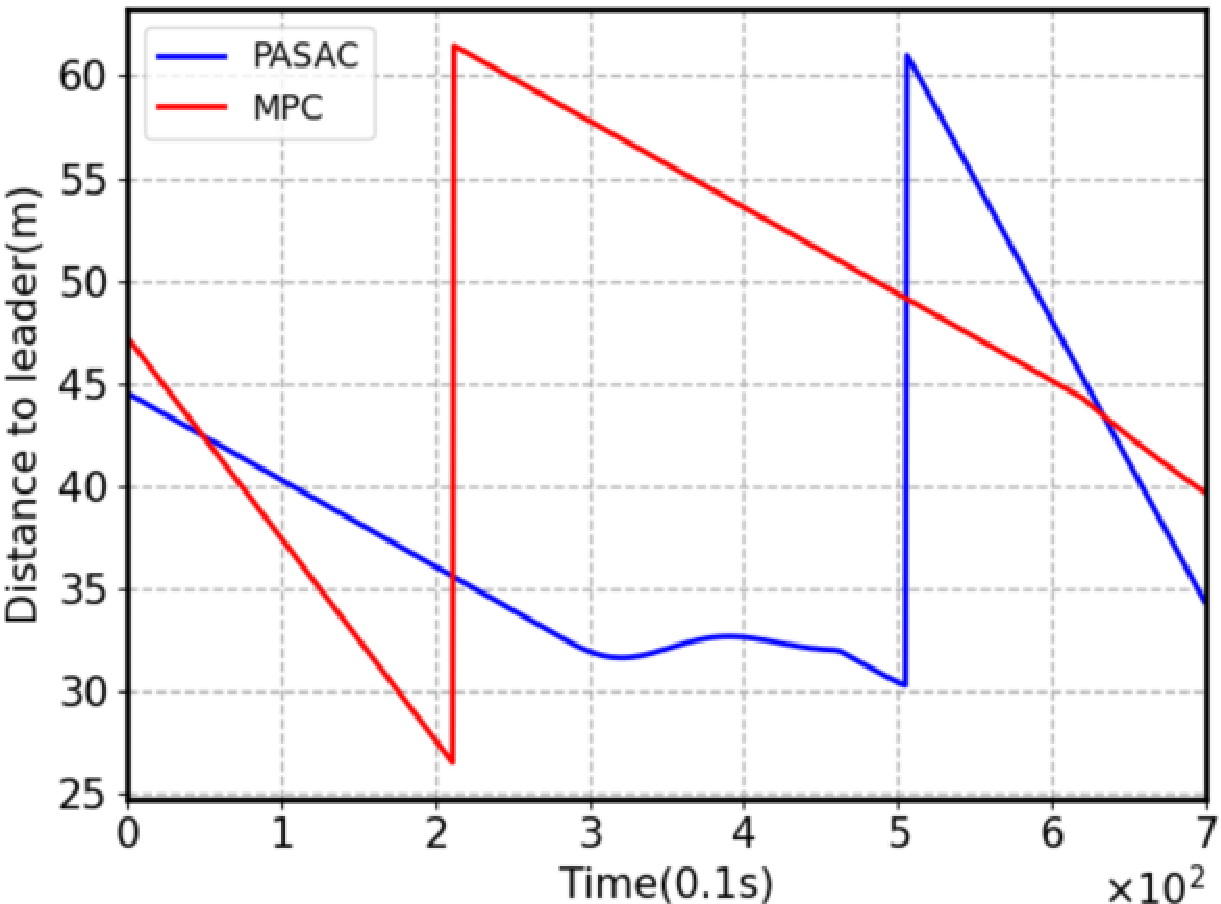} % 更换为图片1的路径
	\includegraphics[width=0.5\textwidth]{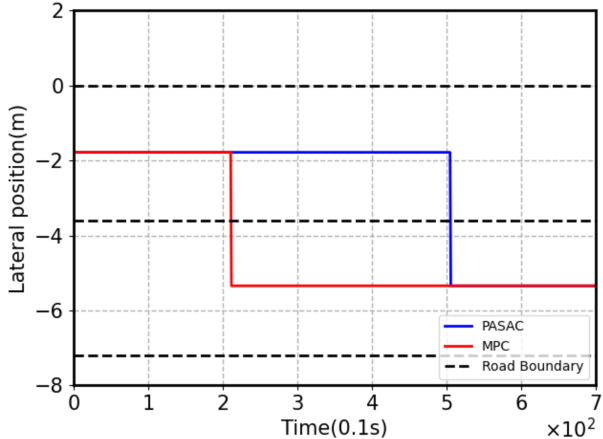} % 更换为图片2的路径
	\caption{Distance to the leader and lateral position during the~simulation. \label{fig6}}
	\label{fig:combinede}
\end{figure}

In the simulation results depicted in Figure~\ref{fig5}, the~acceleration and jerk of both MPC and PASAC demonstrate smooth motion characteristics, avoiding abrupt accelerations and vibrations, significantly enhancing the comfort of the driver. Figures~\ref{fig6} 
 depicts the lateral position and distance to the leading vehicle in the simulations of PASAC and MPC for the ego~vehicle.

\vspace{-9pt}

Figure~\ref{fig6} illustrates the distance from the ego vehicle to the leading vehicle, with~a set safety following distance of 25 m. To~optimize the safety, both MPC and PASAC choose to initiate lane changes before reaching the 25 m distance to the leading vehicle. Note that there was a sudden change in the distance between the ego vehicle and the leading vehicle, indicating a successful lane change and a subsequent alteration in the state of the leading~vehicle.

 The lateral position of the autonomous vehicle on the road is shown in Figure~\ref{fig6}, with~dashed lines representing the road boundaries and solid lines distinctly outlining the precise location of the vehicle along the center line of the road. Through the visual contrast between the dashed and solid lines, we can clearly observe where the ego vehicle initiated lane~changes.

\subsection{Generalization~Analysis}
The complexity and dynamism of the environment can lead to a sub-optimal performance of the policy obtained during the training phase when applied to unseen settings. To~thoroughly assess the algorithm's performance, we conducted a series of tests encompassing 100 episodes, including traffic densities of 0.05 and 0.20. Table~\ref{tab4} presents the performance metrics, including the collision rate, average speed, and~cost, across the various testing stages. Our findings indicate that, following testing in multiple traffic densities, the~algorithm exhibited relatively stable performance in new~environments.
\begin{table}[H]
	\caption{The generalization results across different traffic~densities for 100 episodes.\label{tab4}}
	\begin{adjustwidth}{-\extralength}{0cm}
		\newcolumntype{C}{>{\centering\arraybackslash}X}
		\begin{tabularx}{\fulllength}{CccCCCC}
			\toprule
			& 	&   	\textbf{Collision} & \textbf{Average Speed (m/s)}  &\textbf{Lane Change Times} & \textbf{Reward (Cost)}&\textbf{Reward (Cost) Difference}  \\
			\midrule
			\multirow[m]{2}{*}{{\shortstack{ Traffic flow density\\${\phi=0.05}$~(veh/s %MDPI: We removed italic of unit, please confirm.
)}}}
			&PASAC	& 0\%	& 14.40&24	& $-$25.74 &29.78\% \\
		    &MPC  &0\%  &13.97 & 19  & $-$36.66 &0\% \\
			\midrule
			\multirow[m]{2}{*}{{\shortstack{Traffic flow density\\${\phi=0.20}$~(veh/s)}}}
			&PASAC	& \textbf {0.2\% %MDPI: Please confirm if the bold is unnecessary and can be removed. The following highlights are the same.
}	& 14.25&46	& $-$27.53&30.63\% \\
            &MPC  &0\%  &13.92 &33 & $-$39.69&0\%  \\
			\bottomrule
		\end{tabularx}
	\end{adjustwidth}
	\noindent{\footnotesize{ The traffic flow density is $\phi$ = 0.11 vehicles/second (veh/s) during training.}}
\end{table}

\vspace{-15pt}
\section{Conclusions}
In this study, we used a hybrid-action reinforcement learning algorithm, PASAC, and~compared it with MPC for decision and control problems of autonomous vehicles during the lane-change process. Both MPC and PASAC achieved a collision rate of 0\%. They shared the same control update frequency and were capable of handling hybrid-action space problems. We maintained identical testing conditions for PASAC and MPC, including traffic density and traffic scenarios. The~results indicated that, in~the absence of modeling errors, PASAC outperformed MPC in terms of the value function. Nevertheless, the~PASAC algorithm still encountered collisions in scenarios with higher traffic flow, due to inadequate machine learning generalization. One of the challenges lies in the lack of theoretical analysis of the relationship between neural networks and optimal control, which could be a crucial area for future research.In the future, we also will consider more complex conditions, such as harsh weather conditions and unexpected road~incidents.    

%%%%%%%%%%%%%%%%%%%%%%%%%%%%%%%%%%%%%%%%%%
%\section{Patents}

%This section is not mandatory, but may be added if there are patents resulting from the work reported in this manuscript.

%%%%%%%%%%%%%%%%%%%%%%%%%%%%%%%%%%%%%%%%%%
\vspace{6pt} 

%%%%%%%%%%%%%%%%%%%%%%%%%%%%%%%%%%%%%%%%%%
%% optional
%\supplementary{The following supporting information can be downloaded at:  \linksupplementary{s1}, Figure S1: title; Table S1: title; Video S1: title.}


%%%%%%%%%%%%%%%%%%%%%%%%%%%%%%%%%%%%%%%%%%
\authorcontributions{Conceptualization, Y.L.; methodology, X.L., Z.Z., and~Y.L.; formal analysis, X.L., Y.L., and~Z.Z.; investigation, X.L. and Y.L.; data curation, X.L.; writing---original draft preparation, X.L.; writing---review and editing, Y.L. and X.L.; supervision, Y.L. All authors have read and agreed to the published version of the~manuscript. }

\funding{This work was supported in part by Guangzhou Basic and Applied Basic Research Program under Grant 2023A04J1688, and in part by South China University of Technology faculty start-up fund. %MDPI: we revised the content of funding, please confirm. If you have any funding support, please add it here. 
}

\institutionalreview{Not applicable.}

\informedconsent{Not applicable.}

\dataavailability{The data can be obtained upon reasonable request from the corresponding author.} 

\conflictsofinterest{The authors declare that they have no known competing financial interest or
	personal relationships that could have appeared to influence the work reported in this~paper.} 

%%%%%%%%%%%%%%%%%%%%%%%%%%%%%%%%%%%%%%%%%%

%%%%%%%%%%%%%%%%%%%%%%%%%%%%%%%%%%%%%%%%%%
\begin{adjustwidth}{-\extralength}{0cm}

\reftitle{References}

\PublishersNote{}
\end{adjustwidth}
\end{document}